\documentclass[runningheads]{llncs}
\usepackage[T1]{fontenc}
\usepackage{graphicx}
\usepackage{subcaption}
\usepackage{amsmath, amsfonts, amssymb}
\usepackage{hyperref, cleveref}
\usepackage{algorithm, algorithmic}
\usepackage{booktabs}
\usepackage{multirow}
\usepackage[table, dvipsnames]{xcolor}
\usepackage{balance}
\newtheorem{assumption}[theorem]{Assumption}
\begin{document}
\title{Is Adversarial Training with Compressed Datasets Effective?}
%
%
\author{Tong Chen \and
Raghavendra Selvan}
\authorrunning{Chen \& Selvan}
%
\institute{Department of Computer Science, \\ University of Copenhagen, \\ Copenhagen, Denmark \\
\email{\{toch,raghav\}@di.ku.dk}}
\maketitle              

\begin{abstract}
Dataset Condensation (DC) refers to the recent class of dataset compression methods that generate a smaller, synthetic, dataset from a larger dataset. This synthetic dataset aims to retain the essential information of the original dataset, enabling models trained on it to achieve performance levels comparable to those trained on the full dataset. Most current DC methods have mainly concerned with achieving high test performance with limited data budget, and have not directly addressed the question of adversarial robustness. In this work, we investigate the impact of adversarial robustness on models trained with compressed datasets. We show that the compressed datasets obtained from DC methods are not effective in transferring adversarial robustness to models. As a solution to improve dataset compression efficiency and adversarial robustness simultaneously, we present a robustness-aware dataset compression method based on finding the Minimal Finite Covering (MFC) of the dataset. The proposed method is (1) provably robust by minimizing the generalized adversarial loss, (2) more effective than DC methods when applying adversarial training over MFC, (3) obtained by a one-time computation and is applicable for any model. \footnote{Source code is available at \url{https://github.com/saintslab/pytoch}.}

\end{abstract}

\section{Introduction} \label{intro}
Scaling up of model sizes and datasets has been important to make some of the recent breakthroughs in Deep Learning (DL) ~\cite{sevilla2022compute}. The energy consumption, and the corresponding carbon footprint, of using these large model sizes and datasets have become a growing concern within the community~\cite{strubell2020energy}. Reducing the energy consumption of DL has been primarily addressed  by improving the resource efficiency of DL models. For instance, using a smaller model to condense a larger model with knowledge distillation has shown to be quite effective in reducing the energy consumption at inference~\cite{hinton2015distilling}. Similarly, the recent class of methods under the umbrella term of {\em dataset distillation} or {\em dataset condensation} (DC) has attempted to condense large datasets into smaller, synthetic, datasets ~\cite{Dataset2021Zhao}. These DC methods incur a one-time cost for dataset compression, which can then be used to train multiple models or perform iterative optimization efficiently~\cite{DCBENCH2022Cui}.

While the trade-off between test performance and resource efficiency is deliberated in the DC literature, there are few works investigating the effects of dataset compression on other aspects of model performance. Consider robustness as a case in point; DL models are known to be vulnerable to tiny perturbations of inputs which was first investigated using adversarial examples~\cite{Explaining2015Goodfellow}. The three-way interaction between test performance, efficiency and adversarial robustness is not fully understood.

In this work, we study the trade-off between the efficiency obtained due to the use of compressed datasets and their impact on the robustness of models by answering the questions: 
\begin{enumerate}
    \item {\em Is Adversarial Training with Compressed Datasets Effective?}
    \item {\em Can we improve data efficiency and robustness simultaneously?} 
\end{enumerate}

To answer these questions we investigate the adversarial robustness of models trained with compressed datasets obtained from several coreset selection and DC methods. Using empirical evidence we show that compressed datasets do not offer a good trade-off between test performance and adversarial robustness. Based on our observations, we formulate a novel, {\em robustness-aware dataset compression} method with provable guarantees, which ensures that adversarial training with the compressed datasets is more effective than with DC methods. We present evidence for this paradigm to work for for low-dimensional data.

\section{Related Works} \label{relate}

{\bf Adversarial robustness}: With the emergence of adversarial attacks \cite{Explaining2015Goodfellow}, understanding and enhancing adversarial robustness have become challenging open problems. Numerous strategies for neural networks to defend against adversarial attacks have been proposed in the last decade, including randomized smoothing \cite{Certified2019Cohen}, adversarial training \cite{Ensemble2018Tramer}, and gradient norm regularization \cite{Scaleable2021Finlay}. RobustBench, a benchmarking platform, offers a comprehensive overview of various adversarial training methods tailored to diverse datasets and norm specifications ($\ell_{\infty}, \ell_2$, etc.)~\cite{Robustbench2020Croce}. 

While adversarial training has proven crucial for enhancing robustness, it only provides an intuitive understanding of downstream performance. Consequently, the research community is also seeking methods that can provide provable guarantees of robustness. These works are usually referred to as \emph{certified training}, which are usually more computationally expensive than adversarial training, but hold particular significance in specific applications. Popular works include interval bound propagation \cite{Effectiveness2018Gowal}, linear relaxation \cite{CROWN2018Zhang}, and diffusion de-noised smoothing \cite{carlini2023certified}.

Note, however, adversarial robustness is achieved at the expense of test performance~\cite{Robustness2019Tsipras} and increased computational costs~\cite{Shafahi2019Adversarial}.

{\bf Robustness-aware model compression:}
The computational expense associated with solving non-convex min-max problems, particularly given the increasing need to scale up to large datasets, poses challenges for adversarial training. Currently, the main methods of improving the efficiency of adversarial robustness are by using compressed models, as shown in works such as adversarial robust distillation~\cite{Adversarially2020Goldblum} and robust model compression~\cite{Adversarial2019Ye}. These methods are based on jointly training a smaller model while preserving the adversarial robustness. Recent works have also investigated robust-fine tuning of compressed models~\cite{thorsteinsson2024adversarial}.

{\bf Dataset compression: } At a high-level dataset compression methods can be categorized into two: coreset selection, and DC. Coreset selection methods select a subset of representative samples from the original dataset and include classical techniques such as K-center selection~\cite{farahani2009facility} and herding~\cite{Coresets2009Chen}. 

DC methods are used to generate a smaller set of synthetic data from the larger original dataset so that a model trained on the smaller dataset yields the same performance as when trained with the original dataset. The original DC method introduced in~\cite{wang2018dataset} synthesized new data by matching performance to a network trained with the original dataset. Further methods have been developed for DC by matching gradients~\cite{Dataset2021Zhao}, training trajectories~\cite{TM2022George}, and intermediate feature-maps~\cite{CAFE2022Wang}. While these DC methods match some form of performance based on some explicit downstream task,  distribution matching (DM) introduced in~\cite{DM2023Zhao} performs DC by matching the statistics of the original and the synthetic distributions in a learnt embedding space (which implicitly could correspond to the downstream task). We refer to ~\cite{DCBENCH2022Cui} for an overview of more recent methods along with benchmarking.

\section{Formalizing the Trade-off between Accuracy and Robustness} \label{sec:tradeoff}
Before going into the experimental results, we discuss some well-known results of accuracy and robustness of models, and how they behave for different datasets. We first give the formal definition of test accuracy and robust accuracy, which are the core concepts throughout the following discussion.

\begin{definition}
    Given a probability measure $\nu$, an i.i.d. sampling $\mathcal{T}$ from $\nu$ with its associated empirical measure $\hat{\nu}_{\mathcal{T}}$, and a hypothesis space $\mathcal{H}$. For $f \in \mathcal{H}$,

    (1) The \emph{standard accuracy} of $f$ is defined as $A^{std} (f) := \mathbb{P}_{(\mathbf{x}, y) \sim \nu} [f (\mathbf{x}) = y]$, with its empirical estimation $\hat{A}^{std} (f, \mathcal{T}) := \mathbb{P}_{(\mathbf{x}, y) \sim \hat{\nu}_{\mathcal{T}}} [f (\mathbf{x}) = y]$;
    
    (2) The \emph{robust accuracy} of $f$ w.r.t. perturbation $\varepsilon$ is defined as $A^{rob} (\varepsilon, f) := \mathbb{P}_{(\mathbf{x}, y) \sim \nu} [f (\mathbf{x} + \delta) = y, \; \forall \; \|\delta\| \le \varepsilon]$, with its empirical estimation $\hat{A}^{rob} (\varepsilon, f, \mathcal{T}) := \mathbb{P}_{(\mathbf{x}, y) \sim \hat{\nu}_{\mathcal{T}}} [f (\mathbf{x} + \delta) = y, \; \forall \; \|\delta\| \le \varepsilon]$.
\end{definition}

It is well known that there is a trade-off between test accuracy and robustness \cite{Robustness2019Tsipras,Theoretically2019Zhang} for deep neural networks. This is to say that, if the standard accuracy of a model $A^{std} (f)$ is high (which can be achieved by standard training), then its robust accuracy $A^{rob} (\varepsilon, f)$ is usually low, i.e., any classifier that is very accurate will necessarily be non-robust. Identically, if the robust accuracy of a model $A^{rob} (\varepsilon, f)$ is high (which can be achieved by adversarial training), then its standard accuracy $A^{std} (f)$ is usually low i.e.:

\begin{center}
    \emph{It is usually difficult for a classifier to be both accurate and robust.}
\end{center}

The accuracy-robustness trade-off we described above is about the performance of classifiers in $\mathcal{H}$ over a fixed probability measure $\nu$ or $\hat{\nu}_{\mathcal{T}}$. For every dataset $\mathcal{T}$, we define the \emph{standard classifier} as $f^{std}_{\mathcal{T}} = \arg \min_{f \in \mathcal{H}} \hat{A}^{std} (f, \mathcal{T})$, and the \emph{robust classifier} as $f^{adv}_{\varepsilon, \mathcal{T}} = \arg \min_{f \in \mathcal{H}} \hat{A}^{rob} (\varepsilon, f, \mathcal{T})$.  If we focus on the optimal classifiers $f^{std}_{\mathcal{T}}, f^{adv}_{\varepsilon, \mathcal{T}}$ depending only on $\mathcal{T}$, then we can also discuss about the accuracy and robustness of different datasets $\mathcal{T}$.
\begin{definition} \label{def:score}
    Given a finite dataset $\mathcal{T}$.
    
    (1) The \emph{standard score} of $\mathcal{T}$ is defined as the test accuracy of the standard classifier $f^{std}_{\mathcal{T}}$, i.e., $S^{std} (\mathcal{T}) = A^{std} (f^{std}_{\mathcal{T}})$;

    (2) The \emph{robust score} of $\mathcal{T}$ w.r.t. perturbation $\varepsilon$ is defined as the robust accuracy of the robust classifier $f^{adv}_{\varepsilon, \mathcal{T}}$, i.e., $S^{rob} (\varepsilon, \mathcal{T}) = A^{rob} (\varepsilon, f^{adv}_{\varepsilon, \mathcal{T}})$.
\end{definition}

In the current DC literature, most of the methods are proposed to maximize the test score of dataset $\mathcal{T}$, resulting in solving the optimization problem: $\max_{\mathcal{T}} S^{std} (\mathcal{T})$. This has shown excellent performance even for extreme dataset compression, despite the expensive computation and heavy memory burden~\cite{DCBENCH2022Cui}. Take the CIFAR10 dataset, for example, we are able to achieve $71.6$\% test accuracy over a synthetic dataset with only $50$ images per class v.s. $84.8$\% over the original dataset with $50k$ images in total, according to \cite{TM2022George}.

However, to the best of the authors' knowledge, there are no existing works concerned about the robustness of models trained with these synthetic, compressed, datasets. In this work, we re-observed the accuracy-robustness trade-off for compressed datasets: if the test score of a dataset $S^{std} (\mathcal{T})$ is high (which can be achieved by DC mehtods), then their robust score $S^{rob} (\varepsilon, \mathcal{T})$ is usually low, which means adversarial training over $\mathcal{T}$ is not effective. We therefore propose the following conjecture:

\begin{center}
    \emph{It is usually difficult for a compressed dataset to be both accurate and robust.}
\end{center}

Similar to \cite{Robustness2019Tsipras}, we give an explicit example to illustrate the above conjecture in Appendix~\ref{app:example}.

\section{Experiments \& Results} 
\label{experiments}
In this section, we provide empirical evidence supporting the trade-off between robustness and accuracy for compressed datasets, as previously discussed. Our experiments are conducted on the following datasets: MNIST, CIFAR10, CIFAR100, SVHN. We explore three DC methods: distribution matching ({DM}) \cite{DM2023Zhao}, gradient matching ({GM}) \cite{Dataset2021Zhao}, trajectory matching ({TM}) \cite{TM2022George}; various coreset methods: minimal coreset (MCS, ours), gradient-based methods {Craig} \cite{mirzasoleiman2020coresets} and {GradM} \cite{killamsetty2021gradmatch}, submodularity based methods with graph cut and facility location functions ({SubMod}) \cite{iyer2021submodular}; and baselines: original full dataset ({Raw}), random coreset selection ({Rand}). The details about our MCS method will be discussed in \cref{sec:ours}.

We use PGD-$\ell_{\infty}$ attack to perform adversarial training and compute the robust accuracy, and use AutoAttack \cite{croce2020autoattack} with APGD-CE and APGD-DLR instead for robustness evaluation. We use ResNet-18 for both standard and adversarial training, the learning rate and number of epochs are fixed at $0.01$ and 20, respectively. Following the common setting in the robustness literature~\cite{Robustbench2020Croce}, we set the adversarial perturbation to $\varepsilon_{\infty} = 0.1$ for MNIST dataset, $\varepsilon_{\infty} = 8/255$ for CIFAR10, CIFAR100, SVHN datasets, for $\ell_{\infty}$-norm. For $\ell_2$-norm, in order to keep the volume of $\ell_{\infty}$-ball and $\ell_2$-ball be similar, we set $\varepsilon_2 = \sqrt{\frac{2n}{\pi e}} \varepsilon_{\infty}$, where $n$ is the (flattened) dimension of input images. We use SGD optimizer in Pytorch\footnote{\url{https://pytorch.org/}} with momentum 0.9 and weight decay $5\times 10^{-4}$, and all the experiments are run with 3 repeats using an NVIDIA A100 40GB GPU. 

\begin{table*}[t]
\vskip -0.5cm
    \setlength{\tabcolsep}{4pt}
    \caption{Downstream performance of models trained over compressed dataset of MNIST, CIFAR10, CIFAR100, and SVHN. The considered coresets are {Craig}, {Forget}~\cite{toneva2018an}, {Glister}~\cite{Killamsetty2020GLISTERGB}, {GradM}, {GraNd}~\cite{paul2021deep}, {Herding}~\cite{welling2009herding}, and {SubMod}. For dataset condensation methods {DM}, {GM}, {TM} are considered. For all methods the budget is fixed size 50. Remind that the standard score and robust score of a dataset is defined by $S^{std} (\mathcal{T}) = A^{std} (f^{std}_{\mathcal{T}})$ and $S^{rob} (\varepsilon, \mathcal{T}) = A^{rob} (\varepsilon, f^{adv}_{\varepsilon, \mathcal{T}})$, according to \cref{def:score}. All robust scores are computed by AutoAttack w.r.t. $\ell_{\infty}$-norm with APGD-CE and APGD-DLR. All experiments are run with 3 repeats and best robust score across all methods is highlighted in grey.} 
    \label{tab:results}
    \begin{center}
        \begin{scriptsize}
            \begin{sc}
                \begin{tabular}{ccccccc}
                \toprule
                {\bf Type} &{\bf Methods} & {\bf Score} & {\bf MNIST} & {\bf CIFAR10} & {\bf CIFAR100} & {\bf SVHN} \\
                \midrule
                \multirow{23}{*}{Coreset} &\multirow{2}{*}{Craig.} & std & 87.02$\pm$0.81&29.24$\pm$2.79&17.46$\pm$2.12&57.37$\pm$1.04 \\
                & & {rob} &72.45$\pm$1.11&\cellcolor{gray!25} 9.01$\pm$1.98 &4.70$\pm$3.22& \cellcolor{gray!25}{19.59$\pm$0.00} \\
                \cmidrule{2-7}
                &\multirow{2}{*}{ Forget} & std &90.62$\pm$1.16&32.14$\pm$0.55 &25.00$\pm$0.34&35.08$\pm$3.28  \\
                && {rob} &75.95$\pm$2.29&8.88$\pm$2.42&7.01$\pm$0.24& 18.45$\pm$1.96\\
                \cmidrule{2-7}
                &\multirow{2}{*}{ Glister} & std &69.06$\pm$0.91&25.96$\pm$0.70&18.94$\pm$0.21&41.35$\pm$2.60 \\
                && {rob} &53.95$\pm$0.77&5.79$\pm$1.42&4.94$\pm$0.17&12.51$\pm$6.20 \\
                \cmidrule{2-7}
                &\multirow{2}{*}{ GradM.} & std &91.24$\pm$0.44 &27.53$\pm$0.47&19.78$\pm$0.35&36.00$\pm$1.18 \\
                && {rob} &86.46$\pm$0.37 &5.83$\pm$1.70&5.24$\pm$0.53&6.69$\pm$1.23 \\
                \cmidrule{2-7}
                &\multirow{2}{*}{ GraNd} & std &54.87$\pm$1.58&19.54$\pm$0.47&13.65$\pm$0.35&16.33$\pm$0.4 \\
                && {rob} &30.15$\pm$5.19 &2.62$\pm$0.46&2.45$\pm$0.14&14.09$\pm$7.62 \\
                \cmidrule{2-7}
                &\multirow{2}{*}{ Herding} & std &62.87$\pm$1.29&31.02$\pm$0.61&14.86$\pm$0.30&15.37$\pm$1.29 \\
                && {rob} &45.39$\pm$1.57 &7.96$\pm$0.41&3.36$\pm$0.24&14.50$\pm$8.81 \\
                \cmidrule{2-7}
                &\multirow{2}{*}{ Submod.} & std & 82.22$\pm$0.02&{38.21$\pm$0.91}&22.89$\pm$0.20 & 53.48$\pm$3.86 \\
                && {rob} &75.40$\pm$0.21&{9.14$\pm$0.26}& \cellcolor{gray!25}{7.35$\pm$0.63}&18.62$\pm$1.66 \\
                \cmidrule{2-7}
                &\multirow{2}{*}{ Rand.} & std & 95.15$\pm$0.14 &32.29$\pm$0.76&20.24$\pm$0.54&65.05$\pm$5.61 \\
                && {rob} & 92.50$\pm$0.37&8.79$\pm$1.05&6.59$\pm$0.23&9.36$\pm$0.58 \\
                \cmidrule{2-7}
                &\multirow{2}{*}{MCS (ours)} & std & 96.42$\pm$0.21 & 24.88$\pm$3.12 & 13.71$\pm$1.05 & 66.61$\pm$5.73 \\
                && {rob} & \cellcolor{gray!25} 92.83$\pm$0.07 & 5.26$\pm$0.16 & 4.84$\pm$0.58 & 13.74$\pm$2.20 \\
                \midrule
                \multirow{7}{*}{Condensation}&\multirow{2}{*}{ DM} & std & 96.70$\pm$0.09 & 32.59$\pm$2.49 & 16.00$\pm$0.95 & 62.23$\pm$13.59 \\
                && {rob} & 91.20$\pm$0.27 & 2.68$\pm$0.30 & 3.52$\pm$0.65 & 4.21$\pm$0.43 \\
                \cmidrule{2-7}
                &\multirow{2}{*}{ GM} & std & 96.54$\pm$0.15 & 30.67$\pm$2.10 & 8.61$\pm$0.88 & 51.68$\pm$13.86 \\
                && {rob} & 83.68$\pm$0.74 & 0.73$\pm$0.21 & 0.99$\pm$0.30 & 0.35$\pm$0.13 \\
                \cmidrule{2-7}
                &\multirow{2}{*}{ TM} & std & 95.93$\pm$0.36 & 30.62$\pm$3.03 &{24.45$\pm$0.14} & 47.08$\pm$10.63 \\
                && {rob} & 82.84$\pm$1.73 & 0.87$\pm$0.19 & 1.09$\pm$0.10 & 1.19$\pm$0.31 \\
                \bottomrule
                \end{tabular}
            \end{sc}
        \end{scriptsize}
    \end{center}
    \vskip -0.5cm
\end{table*}

\Cref{tab:results} summarizes the standard scores and robust scores of models trained on different compressed datasets. The three accuracy-oriented compression methods ({DM}, {GM}, {TM}) demonstrate relatively good test performance but are less effective for adversarial training. For example, on the MNIST dataset, while the {DM}, {GM}, and {TM} methods exhibit excellent test performance (standard scores of 96.70\%, 96.54\%, and 95.93\%, respectively versus 95.15\% for {Rand.}), their robust performance is inferior to that of {Rand.} (91.20\%, 83.68\%, and 82.84\%, respectively, versus 92.50\%). 
Similarly, for the SVHN dataset, which contains digit images from 0 to 9 but with a much larger size, the robust performance of these DC methods is significantly worse than that of {Rand.} (4.21\%, 0.35\%, and 1.19\%, respectively, versus 9.36\%). These experiments provide extensive evidence that DC methods tend to overfit to test performance, while coreset methods, such as random selection, show more balanced behavior in terms of both accuracy and robustness. 

Given these experiments, we are now in a position to answer the first question:
{\em Is Adversarial Training with Compressed Datasets Effective?}

Based on the results in Table~\ref{tab:results} we observe that DC methods have poor adversarial robustness, as they are designed to optimize for test accuracy. Most of the coreset methods fare better than DC methods when it comes to their adversarial robustness. The most striking point from these results is that the baseline random coreset method outperforms all the DC methods based on its robust accuracy, and obtains comparable performance with other coreset methods. 

\section{Bridging the divide between adversarial robustness and dataset compression} \label{sec:ours}
In the previous section, we have seen that adversarial training with compressed datasets might not be all that effective when using DC methods. 
In this section, we will connect coreset and DC methods using a unified formulation by posing these dataset compression methods as instances of minimal finite covering (MFC) methods. In doing so, we will also present our method which allows us to access the {\em minimal coreset} based on MFC that can effectively improve the adversarial robustness with guarantees. 

\subsubsection*{Preliminaries and Notations} \label{pre}
Consider the $n$-dimensional Euclidean space $\mathbb{R}^n$ with norm $\|\cdot\|$. In particular, for $p > 0$, the $\ell_p$-norm over $\mathbb{R}^n$ is defined as $\|\mathbf{x}\|_p = (\sum_{i = 1}^n |x_i|^p)^{1/p}$ if $p < \infty$ and $\|\mathbf{x}\|_p = \max_i |x_i|$ if $p = \infty$. Define a distance metric $d (\mathbf{x}, \mathbf{y}) = \|\mathbf{x} - \mathbf{y}\|$ for any two points $\mathbf{x}$ and $\mathbf{y}$ in $\mathbb{R}^n$. Given a point $\mathbf{x}$ and a dataset $\mathcal{Y}$, the distance between $\mathbf{x}$ and $\mathcal{Y}$ is the infimum of distances between $\mathbf{x}$ and all points in $\mathcal{Y}$, which is given by $d (\mathbf{x}, \mathcal{Y}) := \inf_{\mathbf{y} \in \mathcal{Y}} d (\mathbf{x}, \mathbf{y})$, and the distance between two datasets is defined as $d(\mathcal{X}, \mathcal{Y}) := \inf_{\mathbf{x} \in \mathcal{X}} \inf_{\mathbf{y} \in \mathcal{Y}} d(\mathbf{x}, \mathbf{y})$. The (quasi)-distance from one dataset $\mathcal{X}$ to another dataset $\mathcal{Y}$ is given by $d (\mathcal{X} \to \mathcal{Y}) := \sup_{\mathbf{x} \in \mathcal{X}} d (\mathbf{x}, \mathcal{Y}) = \sup_{\mathbf{x} \in \mathcal{X}} \inf_{\mathbf{y} \in \mathcal{Y}} d (\mathbf{x}, \mathbf{y})$. The \emph{Hausdorff distance} between two datasets $\mathcal{X}$ and $\mathcal{Y}$ is defined as $d_H (\mathcal{X}, \mathcal{Y}) := \max \big\{d (\mathcal{X} \to \mathcal{Y}), \; d (\mathcal{Y} \to \mathcal{X}) \big\}$. 

For $\eta > 0$, the \emph{$\eta$-ball} of $\mathbf{x}$ is defined as $\mathcal{B}_{\eta} (\mathbf{x}) := \{\mathbf{z}: d(\mathbf{x}, \mathbf{z}) \le \eta\}$. The \emph{the $\eta$-fattening} of a dataset $\mathcal{Y}$ is defined as $\mathcal{B}_{\eta} (\mathcal{Y}) := \bigcup_{\mathbf{y} \in \mathcal{Y}} \mathcal{B}_{\eta} (\mathbf{y})$. We say $\mathcal{Y}$ is a \emph{finite covering} of $\mathcal{X}$ with radius $\eta$, or $\mathcal{X}$ is \emph{finitely covered} by $\mathcal{Y}$ with radius $\eta$, if $\mathcal{Y}$ is a finite set and for all $\mathbf{x} \in \mathcal{X}$, there exists $\mathbf{y} \in \mathcal{Y}$ such that $d (\mathbf{x}, \mathbf{y}) \le \eta$, i.e., $\mathcal{X} \subseteq \mathcal{B}_{\eta} (\mathcal{Y})$. Obviously, $\mathcal{X}$ is an $\eta$-finite covering of itself for any $\eta > 0$.

\begin{remark}
    (1) If $d (\mathcal{X} \to \mathcal{Y}) = \eta$, then $\mathcal{Y}$ is an $\eta$-finite covering of $\mathcal{X}$. If $d_H (\mathcal{X}, \mathcal{Y}) = \eta$, then $\mathcal{Y}$ is an $\eta$-finite covering of $\mathcal{X}$, and $\mathcal{X}$ is also an $\eta$-finite covering of $\mathcal{Y}$;
    (2) Any synthetic dataset obtained by coreset and DC methods is a finite covering for some $\eta > 0$.
\end{remark}

\subsubsection*{Minimal Coreset (MCS) Method}
Given a dataset $\mathcal{T} = \{\mathbf{x}_i\}_{i = 1}^N \subseteq \mathbb{R}^n$, the trivial $0$-finite covering is the dataset itself, and a trivial $\infty$-finite covering can be an arbitrary data. Similarly, there always exists a finite covering of $\mathcal{T}$ with arbitrary size. Hence the problem of interest is to find the finite covering with minimum radius ($\eta$) or size ($k$).

\begin{definition} \label{def:mfc}
    For a finite set $\mathcal{S} \subseteq \mathbb{R}^n$.
    
    (1) We say $\mathcal{S}^*$ is an \emph{$\eta$-Minimal Finite Covering ($\eta$-MFC)} of $\mathcal{T}$, if $\mathcal{S}^*$ is a finite covering of $\mathcal{T}$ with fixed radius $\eta$ and minimum size, i.e., $\mathcal{S}$ is one of the minimizers of problem $\min_{\mathcal{S}} \{k: d (\mathcal{T} \to \mathcal{S}) = \eta, \; |\mathcal{S}| = k\};$
    
    (2) We say $\mathcal{S}^*$ is a \emph{$k$-Minimal Finite Covering ($k$-MFC)} of $\mathcal{T}$, if $\mathcal{S}^*$ is a finite covering of $\mathcal{T}$ with fixed size $k$ and minimum radius, i.e., $\mathcal{S}$ is one of the minimizers of problem $\min_{\mathcal{S}} \{\eta: d (\mathcal{T} \to \mathcal{S}) = \eta, \; |\mathcal{S}| = k\}.$
\end{definition}
\begin{proposition} \label{thm:mfc}
    The following statements are equivalent:
    
    (1) $\mathcal{S}^*$ is an $\eta$-MFC of $\mathcal{T}$;

    (2) $\mathcal{S}^* \in \arg \min_{\mathcal{S}} \{k: d_H (\mathcal{T}, \mathcal{S}) = \eta, \; |\mathcal{S}| = k\}$.
\end{proposition}
Proof for this proposition is outlined in Appendix \ref{app:mfc-equiv}.

Notice that, if we assume $\mathcal{S}$ is a subset of $\mathcal{T}$, i.e., a \emph{coreset} of $\mathcal{T}$, then the distance from $\mathcal{S}$ to $\mathcal{T}$ is exactly 0. In this case, the Hausdorff distance between $\mathcal{T}$ and $\mathcal{S}$ reduces to the distance from $\mathcal{T}$ to $\mathcal{S}$, i.e., $d_H (\mathcal{T}, \mathcal{S}) = d (\mathcal{T} \rightarrow \mathcal{S})$, and \Cref{thm:mfc} becomes trivial. The corresponding MFC is called a \emph{minimal coreset (MCS)} of $\mathcal{T}$. Any subset of $\mathcal{T}$ can be described using a binary vector $\mathbf{s} \in [0, 1]^N$, where $s_i = 1$ means $\mathbf{x}_i \in \mathcal{S}$ otherwise $\mathbf{x}_i \notin \mathcal{S}$. In the next theorem, we provide a tractable optimization formulation for MCS.
\begin{theorem} \label{prop:mcs}
    For a given dataset $\mathcal{T} = \{\mathbf{x}_i\}_{i = 1}^N$, define the adjacency matrix of $\mathcal{T}$ w.r.t. radius $\eta$ as 
    $$\mathbf{A} (\eta) := \begin{bmatrix} a_{ij} (\eta) \end{bmatrix}, \; 
    a_{ij} (\eta) = \begin{cases}
        1, & \text{if } d (\mathbf{x}_i, \mathbf{x}_j) \le \eta; \\
        0, & \text{otherwise.}
    \end{cases}$$
    
    (1) The $\eta$-MCS of $\mathcal{T}$ is one of the minimizers of
    \begin{align*}\label{opt:eps-mfc}
        \min_{\mathbf{s} \in \{0, 1\}^N} \; \{\|\mathbf{s}\|_1: \mathbf{A} (\eta) \cdot \mathbf{s} \ge 1\}; \tag{$\eta$-MCS}
    \end{align*}

    (2) The $k$-MCS of $\mathcal{T}$ is one of the minimizers of
    \begin{align*}\label{opt:k-mfc}
        \min_{\eta, \; \mathbf{s} \in \{0, 1\}^N} \; \{\eta: \mathbf{A} (\eta) \cdot \mathbf{s} \ge 1, \; \|\mathbf{s}\|_1 = k\}. \tag{$k$-MCS}
    \end{align*} 
\end{theorem}
The proof for this theorem is outlined in Appendix~\ref{app:mcs}.

\begin{remark}
    (1) If we adapt the terminology from DC literature, then MFC can be considered as a DC method, and MCS as a coreset selection method;
    (2) Finding the $k$-MCS of a given dataset is exactly the \emph{K-center} approach \cite{Active2018Sener} for coreset selection. In particular, the case of $k=1$ is well-known as the \emph{smallest bounding sphere} problem, or the \emph{minimal enclosing ball} problem.
\end{remark}

\begin{figure*}[t]
    \vskip 0.2in
    \begin{center}
        \begin{subfigure}[t]{0.24\textwidth}
            \centering
            \centerline
            {\includegraphics[width=\textwidth]{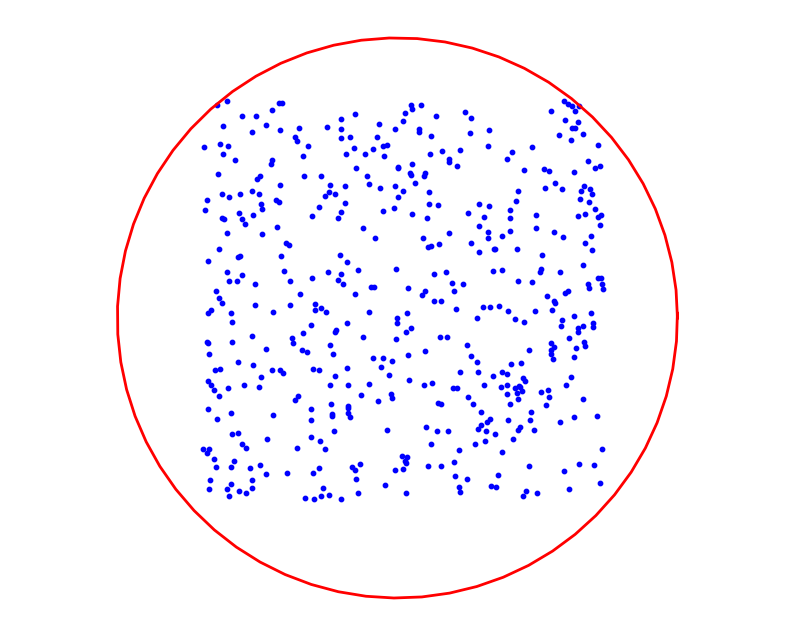}}
            \caption{$\eta = 0.7, k = 1$}
            \label{fig:eta1}
        \end{subfigure}
        \begin{subfigure}[t]{0.24\textwidth}
            \centering
            \centerline
            {\includegraphics[width=\textwidth]{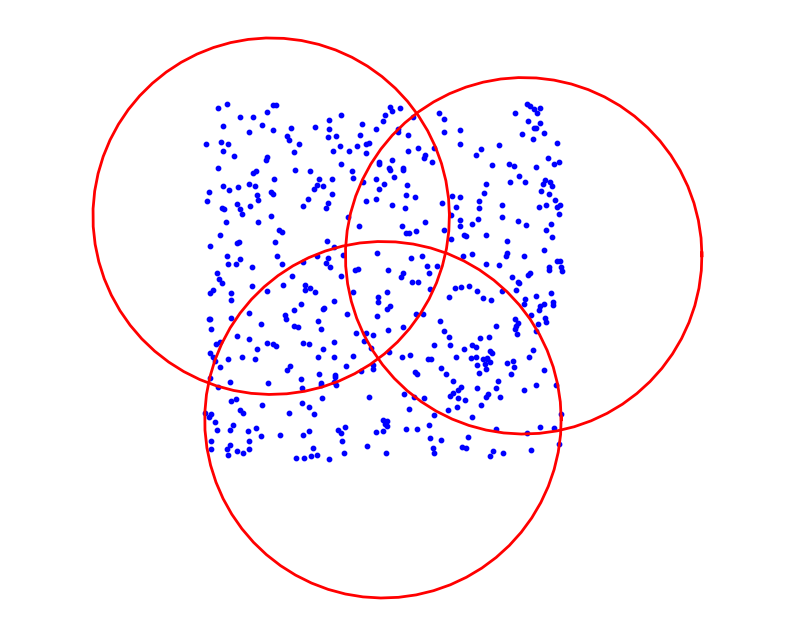}}
            \caption{$\eta = 0.5, k = 3$}
            \label{fig:eta.5}
        \end{subfigure}
        \begin{subfigure}[t]{0.24\textwidth}
            \centering
            \centerline
            {\includegraphics[width=\textwidth]{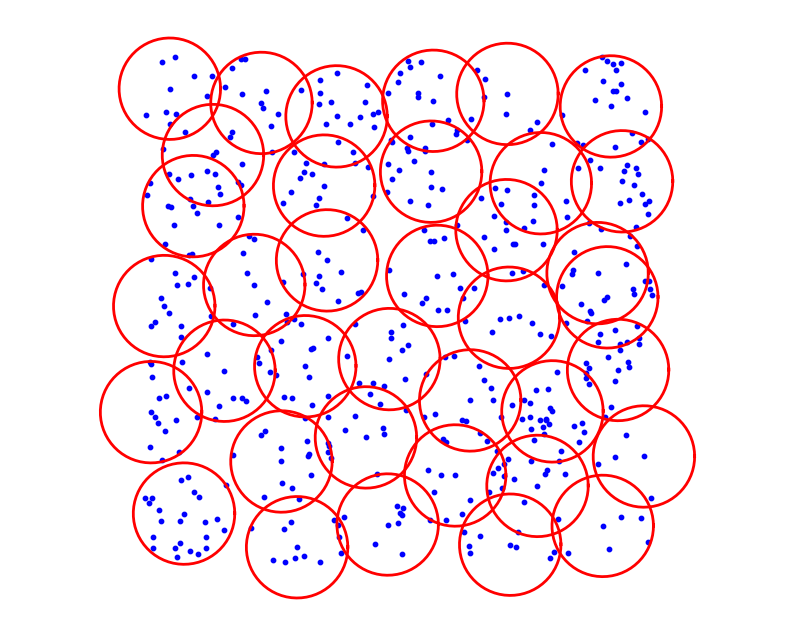}}
            \caption{$\eta = 0.1, k = 37$}
            \label{fig:eta.1}
        \end{subfigure}
        \begin{subfigure}[t]{0.25\textwidth}
            \centering
            \centerline
            {\includegraphics[width=\textwidth]{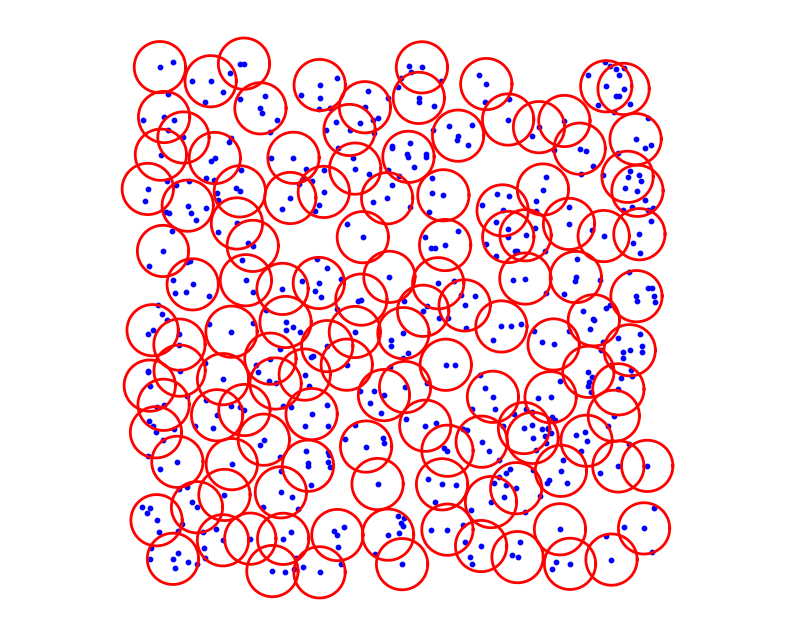}}
            \caption{$\eta = 0.05, k = 127$}
            \label{fig:eta.05}
        \end{subfigure}
        \caption{Visualization of the MFC with fixed radius $\eta$ and the corresponding minimal $k$ for a set of uniformly distributed data points in 2-d. As $\eta$ is reduced, the number of data points required to perform the MFC increase but yield a better covering of the data space. Observe that $\eta=0$ coincides with the original dataset.} \label{fig:icml-dataset}
    \end{center}
    \vskip -0.6cm
\end{figure*}

As we see, determining an $\eta$-MCS is equivalent to solving a \emph{mixed integer linear programming (MILP)} problem, while finding a $k$-MCS involves addressing a \emph{mixed integer quadratically constrained programming (MIQCP)} problem. Since solving a general MILP is already NP-hard, finding a $k$-MCS is much more challenging than finding an $\eta$-MCS. Fortunately, we are able to approach the solution of \eqref{opt:k-mfc} by iteratively solving \eqref{opt:eps-mfc}. The detailed algorithm is presented in \Cref{alg:k-mfc}. Figure~\ref{fig:icml-dataset} illustrates the MFC obtained for 2-dimensional data for different $\eta$ and the corresponding $k$ values using the MCS algorithm.

\begin{algorithm}
    \caption{$k$-MCS} \label{alg:k-mfc}
    \begin{algorithmic}
       \STATE {\bfseries Input:} dataset $\mathcal{T}$, size $k$
       \STATE Initialize $\eta_l \leftarrow 0, \; \eta_u \leftarrow R, \; r \leftarrow \eta_u - \eta_l, \; \delta \leftarrow 10^{-5}$
       \REPEAT
       \STATE $\eta_{new} \leftarrow (\eta_l + \eta_u) / 2$
       \STATE $v \leftarrow$ feasibility of \eqref{opt:k-mfc} with radius $\eta_{new}$
       \IF{$v$ = TRUE}
       \STATE $\eta_l \leftarrow \eta_l, \; \eta_u \leftarrow \eta_{new}$  
       \ELSE
       \STATE $\eta_l \leftarrow \eta_{new}, \; \eta_u \leftarrow \eta_u$  
       \ENDIF
       \STATE $r \leftarrow \eta_u - \eta_l$
       \UNTIL{$r \le \delta$}
       \STATE {\bfseries Output:} MCS with given size $k$ and minimum radius $\eta_u$
    \end{algorithmic}
\end{algorithm}

\subsubsection*{Adversarial Training over $\eta$-MCS}
\label{sec:adv-mcs}

In this section, we will apply standard- and adversarial- training over the MCS we obtained in the previous section, resulting in the robustness-aware dataset compression method.

Given a probability measure $\nu$ and a finite dataset $\mathcal{T} \subseteq \mathbb{R}^{n+1}$, we assume each data point $(\mathbf{x}, y) \in \mathcal{T}$ is i.i.d. sampled from $\nu$. Each finite sampling $\mathcal{T}$ defines an empirical distribution $\hat{\nu}_{\mathcal{T}} = \frac{1}{|\mathcal{T}|} \sum_{(\mathbf{x}, y) \in \mathcal{T}} \delta_{(\mathbf{x}, y)}$, where $\delta_{(\mathbf{x}, y)}$ is the Dirac measure over point $(\mathbf{x}, y)$. For a loss function $l: \mathbb{R}^2 \rightarrow \mathbb{R}_+$, the standard training aims to learn a model $f$ that minimizes the standard loss \\ $L^{std} (f) = \mathbb{E}_{(\mathbf{x}, y) \sim \nu} [l (f (\mathbf{x}), y)]$, with its empirical estimation \\ $\hat{L}^{std} (f, \mathcal{T}) = \mathbb{E}_{(\mathbf{x}, y) \sim \hat{\nu}_{\mathcal{T}}} [l (f (\mathbf{x}), y)]$. On the other hand, adversarial training aims to learn a model that minimizes the adversarial loss w.r.t. some perturbation $\varepsilon > 0$, defined by $L^{adv}_{\varepsilon} (f) = \mathbb{E}_{(\mathbf{x}, y) \sim \nu} [\max_{\|\delta\| \le \varepsilon} l (f (\mathbf{x} + \delta), y)]$, with its empirical estimation $\hat{L}^{adv}_{\varepsilon} (f, \mathcal{T}) = \mathbb{E}_{(\mathbf{x}, y) \sim \hat{\nu}_{\mathcal{T}}} [\max_{\|\delta\| \le \varepsilon} l (f (\mathbf{x} + \delta), y)]$.

In the following discussion, we will focus on classification task, i.e., $y \in \{1, \ldots, K\}$. For a probability measure $\nu$ and a fixed label $i$, we denote by $\mathcal{T}_i \subseteq \mathbb{R}^n$ an i.i.d sampling from the conditional probability measure $\nu (\cdot | y = i)$. We assume that $\mathcal{T}_i \cap \mathcal{T}_j = \varnothing$ for all $i \ne j$, since one single point cannot share two different labels. In adversarial training, we are also searching around the ball of radius $\varepsilon$ for each point, which is exactly the $\varepsilon$-fattening of dataset $\mathcal{T}$. Therefore, we also need to make sure that the two fattening balls of different labels do not overlap.

\begin{definition}
    We say that $\{\mathcal{T}_i\}$ satisfy the \emph{Zero Intersection Property (ZIP)} if for all $i \ne j$, the sets $\mathcal{T}_i$ and $\mathcal{T}_j$ do not overlap, i.e., $\mathcal{T}_i \cap \mathcal{T}_j = \varnothing$.
\end{definition}

\begin{assumption} \label{assump:rip}
    In standard training, we assume $\{\mathcal{T}_i\}$ satisfies ZIP. In adversarial training, we assume $\varepsilon$ is such that $\{\mathcal{B}_{\varepsilon} (\mathcal{T}_i)\}$ satisfies ZIP. In this case, we also say that $\mathcal{T}$ is \emph{$\varepsilon$-separated} borrowing the terminology in \cite{ACloser2020Yang}.
\end{assumption}
\begin{remark}
    \Cref{assump:rip} is easily satisfied by setting $\varepsilon < \min_{j \ne k} d(\mathcal{T}_j, \mathcal{T}_k) / 2$.
\end{remark}

While performing adversarial training over MCS, we can simply use the classical definition of adversarial loss, i.e., set equal weights to each individual Dirac measure in the empirical distribution $\hat{\nu}_{\mathcal{T}}$. However, by doing this, we are unable to derive any provable relation between the empirical adversarial loss over the compressed dataset and the original one. Fortunately, for MCS, we can define the generalized adversarial loss to provide a provable guarantee compared to the classical loss.

\begin{definition} \label{def:gadv}
    Let $\mathcal{T} = \cup_i \mathcal{T}_i$ be a finite sampling from $\nu$ where each $\mathcal{T}_i$ is the conditional sampling from $\nu (\cdot | y = i)$. Suppose $\mathcal{S} = \cup_i \mathcal{S}_i$ is an $\eta$-MCS of $\mathcal{T}$ such that $\{\mathcal{B}_{\varepsilon + \eta} (\mathcal{S}_i)\}$ satisfies ZIP,
    
    (1) The \emph{generalized empirical distribution} over $\mathcal{S}$ is defined as \\ $\hat{\mu}_{\mathcal{S}} = \frac{1}{|\mathcal{T}|} \sum_{(\mathbf{x}, y) \in \mathcal{S}} q_{(\mathbf{x}, y)} \delta_{(\mathbf{x}, y)}$, where $q_{(\mathbf{x}, y)}$ is the number of points in $\mathcal{T}$ that belongs to the ball $\mathcal{B}_{\eta} (\mathbf{x})$;
    
    (2) The \emph{generalized adversarial loss} over $\mathcal{S}$ is defined as $\hat{G}^{adv}_{\varepsilon + \eta} (f, \mathcal{S}) := \mathbb{E}_{(\mathbf{x}, y) \sim \hat{\mu}_{\mathcal{S}}} [\max_{\|\delta\| \le \varepsilon + \eta} l (f (\mathbf{x} + \delta), y)]$.
\end{definition}

By adding non-trivial weights to the classical adversarial loss, we are able to derive a provable guarantee of the generalized adversarial loss.
\begin{proposition} \label{thm:main}
    Let $\mathcal{T}$ and $\mathcal{S}$ be as defined in \Cref{def:gadv}. Then $\hat{L}^{adv}_{\varepsilon} (f, \mathcal{T}) \le \hat{G}^{adv}_{\varepsilon + \eta} (f, \mathcal{S})$.
\end{proposition}
Proof for this theorem is outlined in Appendix~\ref{app:main}.
From \Cref{thm:main}, we are glad to see that, minimizing the generalized adversarial loss over MCS provides a valid upper bound of minimizing the classical adversarial loss over the original dataset. Note that if $\eta=0$, the generalized adversarial loss reduces to the classical one.

\begin{corollary}
    Let $\mathcal{T}$ and $\mathcal{S}$ be as defined in \Cref{def:gadv}, then the robust model obtained by minimizing the generalized adversarial loss $\hat{G}^{adv}_{\varepsilon + \eta} (f, \mathcal{S})$ is robust over the original dataset $\mathcal{T}$ w.r.t. radius $\varepsilon$.
\end{corollary}

\vspace{-0.25cm}
\subsubsection*{Experiments with MCS}
\vspace{-0.25cm}

The theoretical results that culminate with the main result in~\Cref{thm:main} indicate that the MCS method can cover the data space using MFC while also ensuring adversarial robustness under certain conditions (ZIP in~\Cref{assump:rip}). We now report these results on MNIST dataset using ResNet-18 and AutoAttack for different coreset methods (Rand., SubMod., Craig, Forget, GradM) and DC methods (TM, GM, DM), along with the scores for full dataset. All the experiments are run with 3 repeats.

\begin{figure}[t]
    \centering
    \includegraphics[width=0.75\linewidth]{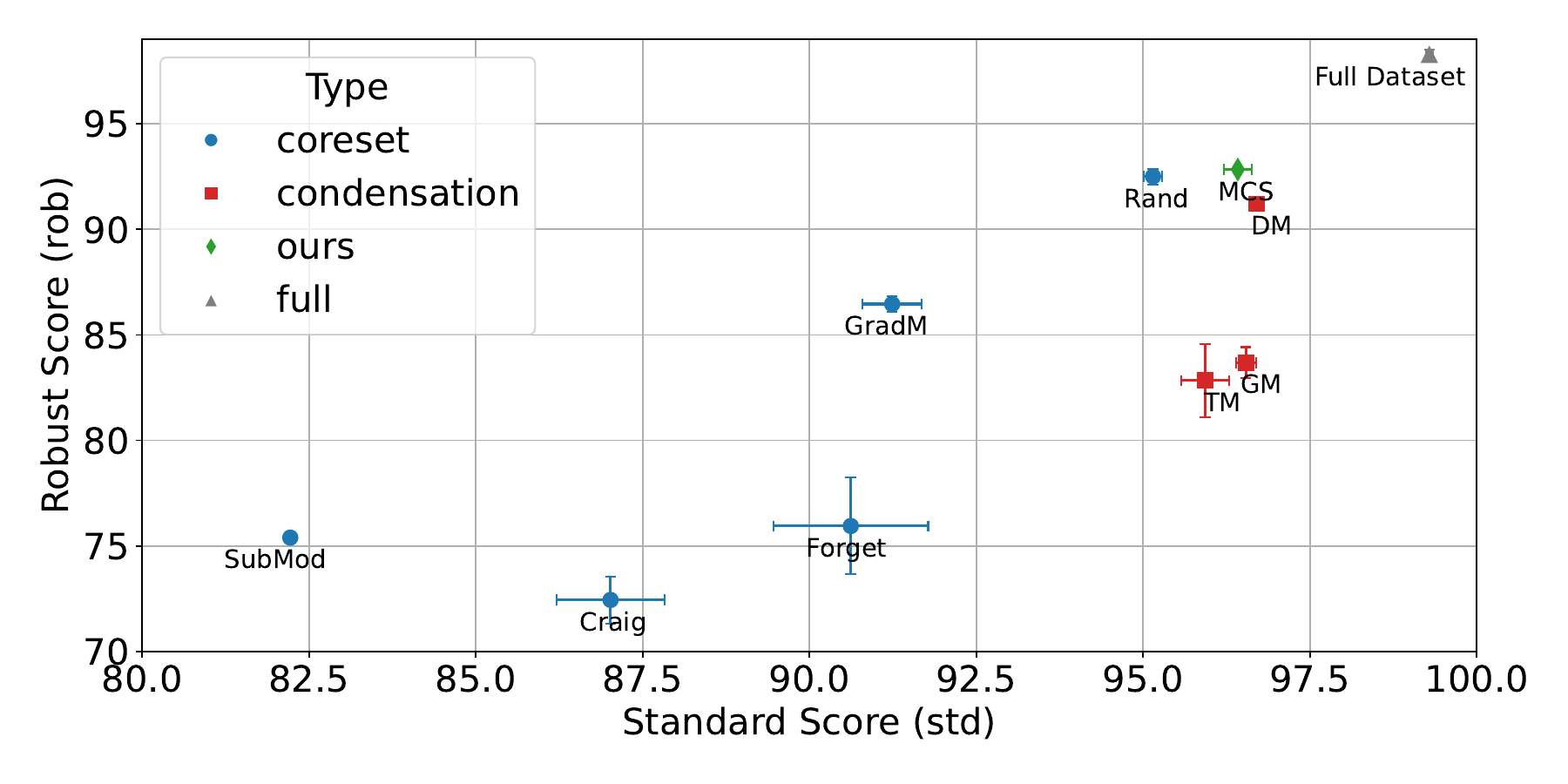}
    \vspace{-0.25cm}
    \caption{Downstream performance of models trained over full and compressed dataset of MNIST. For all methods the budget is fixed size 50. All robust scores are computed by AutoAttack w.r.t. $\ell_{\infty}$-norm with APGD-CE and APGD-DLR. All error bars are based on 3 random repeats.}
    \label{fig:mcs}
    \vspace{-0.5cm}
\end{figure}

The scatter plot in Figure~\ref{fig:mcs} shows the standard and robust accuracy of these methods on MNIST dataset for a budget of 50. Our proposed method MCS with k=50 is also shown. The scatter plot reveals clear distinctions between coreset and condensation methods in terms of both standard and robust accuracy. Condensation methods, such as DM, GM, and TM, consistently achieve high accuracy across both metrics, with DM standing out as one of the most balanced performers, reaching a robust accuracy of 91.20\%.  

Among the coreset methods, MCS emerges as the best performer, surpassing all other methods with a standard accuracy of 96.42\% and a robust accuracy of 92.83\%. This suggests that MCS is not only effective in maintaining overall accuracy but also in preserving robustness. Other coreset methods, such as GradM and Forget, also achieve reasonable robust accuracy, though they do not reach the levels of the condensation methods.  

Overall, MCS proves that coreset methods can also achieve high performance, challenging the notion that condensation methods are always superior, while providing theoretical guarantees on robustness as shown in \Cref{thm:main}. This answers the second question we set out to investigate: {\em Can we improve data efficiency and robustness simultaneously?} 

\vspace{-0.25cm}
\section{Discussion and Conclusion} \label{conclusion}
\vspace{-0.25cm}

The experimental results presented in  Section~\ref{experiments} show DC methods like DM, GM, TM achieve slightly higher test performance in specific instances but have
poor adversarial performance. MCS achieves better adversarial robustness compared to all DC methods. These results further strengthen the key conclusion in our work: {\em The efficacy of  adversarial training diminishes as compressed datasets increasingly overfit to optimize test performance.}

The performance of MCS in some settings is still lower than Random sampling, which is to be expected in low budget regimes (here with 50 samples). This is expected since many of the coreset methods performs worse than random sampling in terms of test performance, according to literature such as~\cite{Dataset2021Zhao}. However, we would like to emphasize that, downstream performance is just one aspect of our method. Moreover, MCS can be regarded as the deterministic version of random sampling, since both of them converges to the full dataset as the budget increases. 

Note that the assumptions to satisfy ZIP in Section~\ref{sec:adv-mcs} are not satisfied by complex datasets. Proposition~\ref{thm:mfc} suggests that discovering an MFC is equivalent to minimizing the Hausdorff distance, ensuring a precise and deterministic covering of the dataset. However, the ZIP condition is related to downstream property but not during the dataset compression phase. The MCS approximates the distribution of the entire dataset, which does not directly impact downstream performance. 

We have the option to ease the constraints of the Hausdorff distance by substituting the maximum with the mean of all distance pairs,
\begin{equation*}
    d^\prime_H (\mathcal{X}, \mathcal{Y}) = \frac{1}{2}\left( \frac{1}{|\mathcal{X}|} \sum_{\mathbf{x} \in \mathcal{X}} \min_{\mathbf{y}\in\mathcal{Y}} d(\mathbf{x},\mathbf{y}) + \frac{1}{|\mathcal{Y}|} \sum_{\mathbf{y} \in \mathcal{Y}} \min_{\mathbf{x}\in\mathcal{X}} d(\mathbf{x},\mathbf{y})\right)
\end{equation*}
Despite, $d^\prime_H (\mathcal{X}, \mathcal{Y})$ loses its property as a distance metric, it remains feasible to derive a coreset that minimizes the relaxed Hausdorff distance. This could yield a soft-margin adaptation of the proposed method.

\textbf{Limitations}: In the first part of the paper, we have benchmarked diffferent coreset and DC methods showing their adversarial robustness. This benchmarking could be extended to other -- more recent DC -- methods and to include additional attacks. We only demonstrated the usefulness of MCS for low-dimensional data (MNIST). Extending these experiments to other non-ZIP satisfying datasets will be important. The proposed MCS method is model-free and is more general purpose as it is not optimised for a specific downstream task. A customized version of finite covering tailored for particular tasks would be more beneficial and interesting in certain application scenarios. Additionally, the choice of distance metric, beyond Euclidean distance, might impact downstream performance, which is not sufficiently discussed in the current paper.

\textbf{Future work}: There are many interesting future directions regarding DC and robustness. For example, we showed that compressed datasets obtained by DC methods have poor robustness scores. Investigations into enabling robust DC methods will be of interest. Further proving the accuracy-robustness trade-off in a more general setting will also be interesting. It has also been demonstrated that random selection performs quite well, at least in comparison to the majority of coreset methods, as shown in our investigation. Providing a theoretical explanation for why random selection exhibits such favorable behavior is crucial and would be an interesting avenue for further research. Finally, investigating the joint-improvements of efficiency with other aspects such as fairness (in terms of both test performance and fairness), and privacy (membership inference privacy) are direct extensions of the current work which only focuses on robustness.

\textbf{Conclusion}: In this work, we stress the accuracy-robustness trade-off in terms of compressed datasets, which is rarely discussed in the robustness and DC communities. We claim that improving both the test score and robust score of a dataset is difficult, because DC methods force the dataset to {\em only} maximize the test performance and violates the underlying data distribution. In order to perform adversarial training with compressed datasets with provable guarantee, we propose MCS and generalized adversarial loss over MCS. We proved that the generalized adversarial loss over MCS provides a valid upper bound of the classical adversarial loss over the original dataset. Finally, we propose an MILP formulation to obtain MCS with a fixed radius or size. We present empirical evidence to substantiate the theoretical guarantees derived for MCS on low-dimensional data.

\section*{Acknowledgments and Disclosure of Funding}
TC and RS acknowledge funding received under  European Union’s Horizon Europe Research and Innovation programme under grant agreements No. 101070284, No. 10107040 and No. 101189771. RS also acknowledges funding received under Independent Research Fund Denmark (DFF) under grant 4307-00143B. The authors thank members of \hyperlink{https://saintslab.github.io/}{SAINTS Lab} for useful discussions. 

\bibliographystyle{plain}
\bibliography{main.bib}

\newpage
\onecolumn
\appendix

\section{Proofs}
\subsection{Proof for \Cref{thm:mfc}} \label{app:mfc-equiv}
\begin{proof}
    Suppose $\mathcal{S}^*$ is an $\eta$-MFC of $\mathcal{T}$, and $\mathcal{S}_H^*$ is a minimizer of $\min_{\mathcal{S}} \{k: d_H (\mathcal{T}, \mathcal{S}) = \eta, \; |\mathcal{S}| = k\}$. We only need to prove $|\mathcal{S}^*| = |\mathcal{S}_H^*|$. Indeed, any dataset $\mathcal{S}$ satisfying $d_H (\mathcal{T}, \mathcal{S}) = \eta$ also satisfies $d (\mathcal{T} \rightarrow \mathcal{S}) \le \eta$, hence $|\mathcal{S}^*| \le |\mathcal{S}_H^*|$. If $|\mathcal{S}^*| < |\mathcal{S}_H^*|$, then we have $d (\mathcal{S}^* \rightarrow \mathcal{T}) > \eta$. By definition, there exists $\mathbf{x}^* \in \mathcal{S}^*$ such that $d(\mathbf{x}^*, \mathbf{y}) > \eta$ for all $\mathbf{y} \in \mathcal{T}$. However, this means by removing point $\mathbf{x}^*$ from $\mathcal{S}^*$, the set $\mathcal{S}^* \setminus \{\mathbf{x}^*\}$ is also a finite covering of $\mathcal{T}$ with radius $\eta$ and size $|\mathcal{S}^*|-1$, which contradicts the minimality of $|\mathcal{S}^*|$.
\end{proof}

\subsection{Proof for \Cref{prop:mcs}} \label{app:mcs}
\begin{proof}
    (1) By the first part of \Cref{def:mfc}, an $\eta$-MFC of $\mathcal{T}$ is one of the minimizers of
    \begin{align} \label{eq:eps-mfc}
        \min_{\mathcal{S}} \{k: d (\mathcal{T} \to \mathcal{S}) = \eta, \; |\mathcal{S}| = k\} = & \min_{\mathcal{S}} \{|\mathcal{S}|: \sup_{\mathbf{x} \in \mathcal{T}} \inf_{\mathbf{y} \in \mathcal{S}} d (\mathbf{x}, \mathbf{y}) = \eta\} \notag \\
        = & \min_{\mathcal{S}} \{|\mathcal{S}|: \forall \mathbf{x} \in \mathcal{T}, \; \exists \mathbf{y} \in \mathcal{S}, \; d(\mathbf{x}, \mathbf{y}) \le \eta\} \notag \\
        = & \min_{\mathcal{S}} \{|\mathcal{S}|: \forall \mathbf{x} \in \mathcal{T}, \; |\mathcal{S}_{\mathbf{x}}| \ge 1\}
    \end{align}
    where $\mathcal{S}_{\mathbf{x}} = \{\mathbf{y} \in \mathcal{S}: d(\mathbf{x}, \mathbf{y}) \le \eta\} \subseteq \mathcal{S}$ for $\mathbf{x} \in \mathcal{T}$. Denote by $\mathbf{A}_{\mathcal{T}, \mathcal{S}} (\eta)$ the adjacency matrix
    $$\mathbf{A}_{\mathcal{T}, \mathcal{S}} (\eta) := \begin{bmatrix} a_{\mathbf{x}, \mathbf{y}} (\eta) \end{bmatrix}, \; 
    a_{\mathbf{x}, \mathbf{y}} (\eta) = \begin{cases}
        1, & \text{if } d (\mathbf{x}, \mathbf{y}) \le \eta; \\
        0, & \text{otherwise.}
    \end{cases}$$
    Then \cref{eq:eps-mfc} is equivalent to $\min_{\mathcal{S}} \{|\mathcal{S}|: \mathbf{A}_{\mathcal{T}, \mathcal{S}} (\eta) \cdot \mathbf{1}_{|\mathcal{S}|} \ge 1\}$ where $\mathbf{1}_{|\mathcal{S}|}$ is the vector of all ones of dimension $|\mathcal{S}|$, and the inequality is element-wise. This immediately concludes the proof if $\mathcal{S}$ is a subset of $\mathcal{T}$.

    (2) By the second part of \cref{def:mfc}, a $k$-MFC of $\mathcal{T}$ is one of the minimizers of
    \begin{align} \label{eq:k-mfc}
        & \min_{\mathcal{S}} \{\eta: d (\mathcal{T} \to \mathcal{S}) = \eta, \; |\mathcal{S}| = k\} = \min_{\mathcal{S}} \{\eta: \mathbf{A}_{\mathcal{T}, \mathcal{S}} (\eta) \cdot \mathbf{1}_{|\mathcal{S}|} \ge 1, \; |\mathcal{S}| = k\}
    \end{align}
    where the equality is directly derived from the previous part.
\end{proof}

\subsection{Proof for \Cref{thm:main}} \label{app:main}
\begin{proof}
    Let $\mathcal{T} = \{(\mathbf{x}_i, y_i)\}_{i = 1}^N$ and $\mathcal{S} = \{(\tilde{\mathbf{x}}_j, \tilde{y}_j)\}_{j = 1}^M$. Since $\mathcal{S}$ is an $\eta$-MCS of $\mathcal{T}$, for every $(\mathbf{x}_i, y_i) \in \mathcal{T}$, there exists $(\tilde{\mathbf{x}}_{j}, \tilde{y}_{j}) \in \mathcal{S}$ such that $y_i = \tilde{y}_{j}$ and $\mathbf{x}_i \in \mathcal{B}_{\eta} (\tilde{\mathbf{x}}_{j})$, hence $\mathcal{B}_{\varepsilon} (\mathbf{x}_i) \subseteq \mathcal{B}_{\varepsilon + \eta} (\tilde{\mathbf{x}}_{j})$. For each $j = 1, \ldots, M$, define index set $I_j = \{i: \mathbf{x}_i \in \mathcal{B}_{\eta} (\tilde{\mathbf{x}}_j)\} \subseteq \{1, \ldots, N\}$. By definition of adversarial loss over $\mathcal{T}$,
    \begin{align*}
        \hat{L}^{adv}_{\varepsilon} (\theta, \mathcal{T}) & = \mathbb{E}_{(\mathbf{x}, y) \sim \hat{\nu}_{\mathcal{T}}} \bigg[\max_{\|\delta\| \le \varepsilon} l (f_{\theta} (\mathbf{x} + \delta), y)\bigg] \\
        & = \frac{1}{N} \sum_{i = 1}^N \max_{\|\delta\| \le \varepsilon} l (f_{\theta} (\mathbf{x}_i + \delta), y_i) \\
        & = \frac{1}{N} \sum_{j = 1}^M \sum_{i \in I_j} \max_{\|\delta\| \le \varepsilon} l (f_{\theta} (\mathbf{x}_i + \delta), y_i) \\
        & \le \frac{1}{N} \sum_{j = 1}^M \sum_{i \in I_j} \max_{\|\delta\| \le \varepsilon + \eta} l (f_{\theta} (\tilde{\mathbf{x}}_j + \delta), \tilde{y}_j) \\
        & = \frac{1}{N} \sum_{j = 1}^M |I_j| \cdot \max_{\|\delta\| \le \varepsilon + \eta} l (f_{\theta} (\tilde{\mathbf{x}}_j + \delta), \tilde{y}_j) = \hat{G}^{adv}_{\varepsilon + \eta} (\theta, \mathcal{S}).
    \end{align*}
    We conclude the proof by letting $q_{(\mathbf{x}_j, y_j)} = |I_j|$.
\end{proof}

\section{Robustness-accuracy trade-off for compressed dataset}
\label{app:example}

\begin{example} \label{eg:tradeoff}
    \emph{(Robustness-accuracy trade-off for compressed dataset)}. Consider a binary classification task for data $(\mathbf{x}, y) \in \mathbb{R}^{n+1} \times \{\pm1\}$ sampled from distribution $\nu (p)$ defined as follows:
    \begin{align*}
        & y \sim \mathcal{U} (\{\pm 1\}), \\
        & x_1 = \begin{cases}
        +y, & \text{w.p. } p, \\
        -y, & \text{w.p. } 1-p, \\
        \end{cases}, \\
        & x_2, \ldots, x_{n+1} \overset{i.i.d.}{\sim} \mathcal{U} ([(y - 1) / 2, (y + 1) / 2]).
    \end{align*}
    Let the hypothesis space $\mathcal{H}$ be the space of linear classifiers $\text{sign} (\mathbf{w}^T \mathbf{x})$, and consider $\ell_{\infty}$-adversarial robustness with perturbation $\varepsilon=1$. Then, for any $p \in [0.5, 1]$,
    
    (1) Any classifier $f_a (\mathbf{x}) = \text{sign} (w_2 x_2 + \cdots + w_{n+1} x_{n+1})$ with $w_i > 0$ is perfect, i.e., $A^{std} (f_a) = 100\%$, but has robust accuracy 0\%;
    (2) The most robust classifier is $f_r (\mathbf{x}) = \text{sign} (x_1)$, with standard and robust accuracy both equal to $p$.
\end{example}

\Cref{eg:tradeoff} indicates that, on the one hand, we always have perfect but non-robust classifiers $f_a$ for any distribution $\nu(p)$. On the other hand, the robust accuracy of any classifier cannot be higher than $p$. In this case, DC methods can easily achieve the best test performance. However, the compressed datasets might not follow the original data distribution. For the synthetic datasets sampled from distribution $\nu(p)$ with $p \approx 0.5$, it is impossible to improve the robust accuracy of the classifiers to a satisfactory level. 

\section{Additional Results}

\subsection{Dataset compression in large compression budget regime} 
\label{sec:large}
We mainly consider DM from~\cite{DM2023Zhao} as the baseline DC method due to the considerable computational costs associated with obtaining the compressed datasets from other DC methods involving bi-level optimizations; particularly when scaling up the size, such as with $k=500$. We choose the computationally cheaper DM method,  as the baseline which also shows competitive performance compared to other DC methods~\cite{DCBENCH2022Cui}. For each compressed dataset, we consider a multilayer perceptron (MLP) for MNIST dataset, and convolutional neural networks (ConvNet) for CIFAR10 dataset. Specifically, the MLP architecture consists of two hidden layers, each comprising 128 neurons. The ConvNet architecture includes 3 blocks, each containing 128 filters of size $3 \times 3$, followed by instance norm \cite{Instance2016Ulyanov}, ReLU activation and average pooling layers. We perform generalized adversarial training over the MCS, and classical adversarial training over other compressed dataset.

For convenience of comparison, we consider $k$-MCS for both MNIST and CIFAR10 dataset with $k=[50, 100, 200, 300, 400, 500]$. However, for these considered sizes, their corresponding radius $\eta$ is quite large and the $(\varepsilon + \eta)$-fattening of the MCS does not satisfy RIP. Thus \Cref{thm:main} is not applicable here anymore. Nevertheless, we can still perform generalized adversarial training over MCS, with perturbation $\varepsilon$ instead of $\varepsilon + \eta$. We will see later that it actually shows excellent empirical downstream performance. \Cref{tab:param_training} shows the hyperparameters for standard and adversarial training for MNIST and CIFAR10 dataset.

\begin{table}[ht]
    \caption{Hyperparameters of standard and (generalized) adversarial training over coresets of MNIST and CIFAR10 datasets. The considered coresets are \textbf{Raw}, \textbf{MCS}, \textbf{Rand}, and \textbf{DM}. The parameter $\alpha$ is the step size in PGD attack. We see that the learning rate for MCS is much smaller than others, which is a quite interesting for future work to explain it in theory.} \label{tab:param_training}
    \begin{center}
        \begin{sc}
            \begin{tabular}{ccccc}
            \toprule
            dataset & coreset & std-lr & adv-lr & $\alpha$ \\
            \midrule
            \multirow{2}{*}{MNIST} & Raw, Rand, DM & \multicolumn{2}{c}{$10^{-1}$} & \multirow{2}{*}{$10^{-1}$} \\
            & MCS & $10^{-3}$ & $10^{-4}$ & \\
            \midrule
            \multirow{2}{*}{CIFAR10} & Raw, Rand, DM & \multicolumn{2}{c}{$10^{-2}$} & \multirow{2}{*}{$10^{-2}$} \\
            & MCS & $10^{-4}$ & $10^{-5}$ & \\
            \bottomrule
            \end{tabular}
        \end{sc}
    \end{center}
    \vspace{-0.8cm}
\end{table}

\begin{figure*}[ht]
    \vskip -0.2in
    \begin{center}
    \centerline{\includegraphics[width=\textwidth]{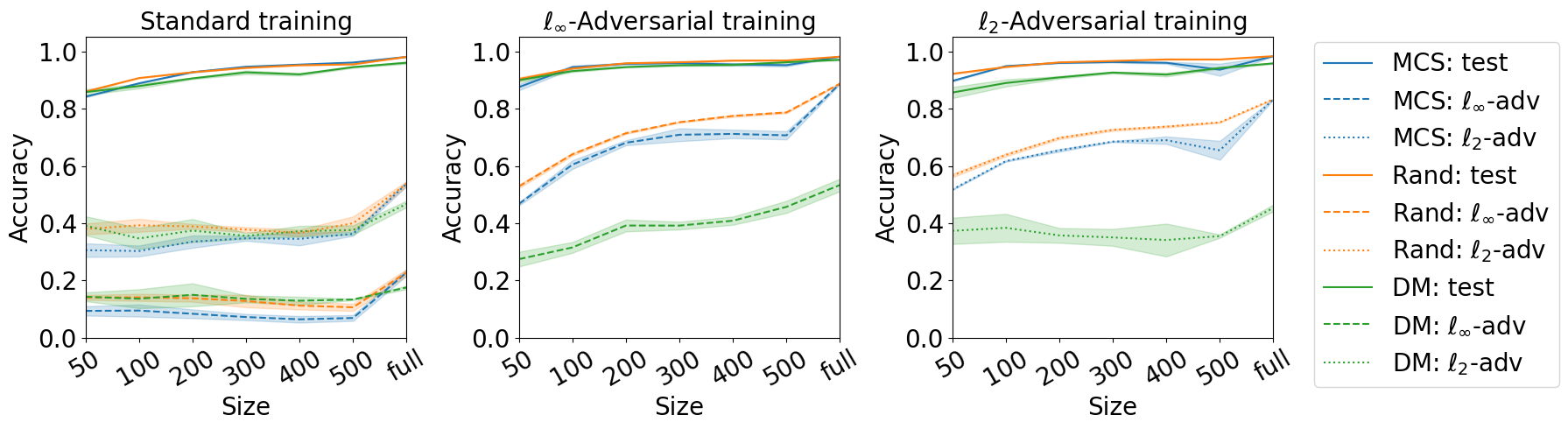}}
        \caption{Performance of standard and robust models trained with different compressed datasets from MNIST. The figures from left to right are standard training, $\ell_{\infty}$-adversarial training and $\ell_2$-adversarial training. The blue, orange and green lines stand for {MCS}, {Rand}, and {DM} respectively. The solid, dashed and dotted lines stand for test, $\ell_{\infty}$ and $\ell_2$ robust accuracy respectively. In the horizontal axis, ``full'' means the original dataset for {MCS} and {Rand}, and size $5000$ for {DM}.} \label{fig:mnist-l2-results}
    \end{center}
    \vskip -0.5in
\end{figure*}

\Cref{fig:mnist-l2-results} shows the test, $\ell_{\infty}$- and $\ell_2$- robust accuracy of models obtained by standard, $\ell_{\infty}$- and $\ell_2$- adversarial training over the original MNIST and compressed dataset. Notice that for {Rand} and {MCS}, the coresets becomes exactly the original dataset if the size equals $N$. However, even though for size $N$, method {DM} is still different from the original dataset. We compute {DM} with size $5000$ for MNIST dataset. We see that, when applying standard training, all models show similar performance in terms of test accuracy and robust accuracy. However, adversarial training over {DM} does not seem to be more effective, while {Rand} and {MCS} have large improvement in robustness. Even if the size of {DM} increases to $5000$, the robust accuracy after adversarial training is still \emph{much} lower than when using the original dataset. This behavior coincides with the conjecture we propose in \Cref{sec:tradeoff}: DC methods violate the underlying distribution of the original data.

\begin{figure*}[ht]
    \begin{center}
    \centerline{\includegraphics[width=\textwidth]{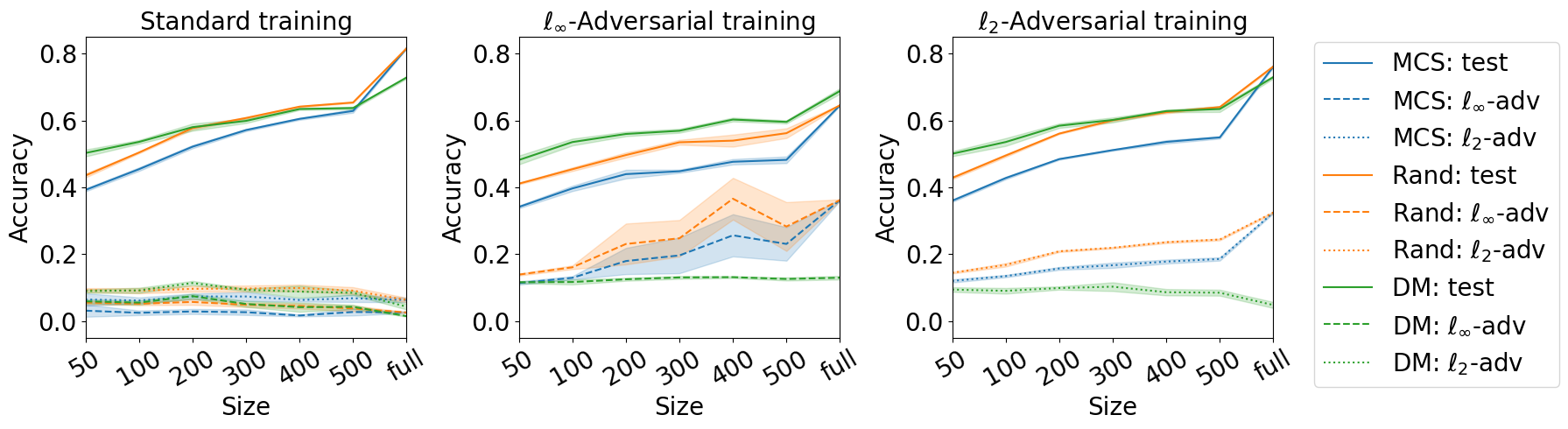}}
        \caption{Performance of standard and robust models trained over different compressed dataset from CIFAR10. The figures from left to right are standard training, $\ell_{\infty}$-adversarial training and $\ell_2$-adversarial training. The blue, orange and green lines stand for {MCS}, {Rand}, and {DM} respectively. The solid, dashed and dotted lines stand for test, $\ell_{\infty}$ and $\ell_2$ robust accuracy respectively. In the horizontal axis, ``full'' means the original dataset for {MCS} and {Rand}, and size $4000$ for {DM}.} \label{fig:cifar10-l2-results}
    \end{center}
    \vskip -0.25in
\end{figure*}

\Cref{fig:cifar10-l2-results} shows exactly the same behavior for CIFAR10 dataset. Due to memory issue, we only have access to {DM} with size $4000$. As the size of compressed dataset increases, the robust accuracy of both {MCS} and {Rand} are both increasing. However, the robustness of {DM} does not seem to improve. Especially if we increase the size $k$ of {DM} to $4000$, the test accuracy of $\ell_{\infty}$-robustly trained model slightly outperforms the original CIFAR10 data (68.86\% v.s. 64.59\%), whereas the robust score of the former is \emph{much} smaller than the latter (12.98\% v.s. 36.21\%), which is actually at the same level of compression size $50$.

\subsection{Standard and robust scores of compressed datasets of MNIST and CIFAR10}
\Cref{tab:summary} provides an overview of all the numerical results evaluated over different models and different datasets. All the results are computed by the mean of 5 repeats and their standard deviation. Surprisingly, the poor behavior of \textbf{DM} is expected, but \textbf{Rand} performs even better than \textbf{MCS} in most cases, both for test and robust accuracy. This makes sense because RIP condition does not hold either for MNIST or CIFAR10, whereas the random  coreset might capture the distribution of the original data better.

\begin{table*}[h]
    \vskip -0.5cm
    \caption{Downstream performance of models trained over original and compressed dataset of MNIST and CIFAR10. The considered coresets are {Raw}, {Rand}-$k$ with $k=500$, and {DM}-$k$ with $k=500$. The $\ell_{\infty}$ perturbation radius of MNIST and CIFAR10 is 0.1 and 8/255 respectively. The $\ell_2$ perturbation radius of MNIST and CIFAR10 is 1.36 and 0.84.} 
    \vskip -0.12in
    \label{tab:summary}
    \begin{center}
        \begin{sc}
            \begin{tabular}{ccccc}
            \toprule
            dataset & method & std. score & $\ell_{\infty}$-score & $\ell_2$-score \\
            \midrule
            \multirow{4}{*}{MNIST} & Raw & 98.08$\pm$0.06 & 88.75$\pm$0.23 & 83.29$\pm$0.19 \\
            \cmidrule{2-5}
            & Rand & 95.47$\pm$0.02 & \textbf{78.69}$\pm$0.28 & \textbf{75.23}$\pm$0.20 \\
            & DM & 94.51$\pm$0.28 & 45.72$\pm$2.11 & 35.50$\pm$0.57 \\
            \midrule
            \multirow{4}{*}{CIFAR10} & Raw & 81.64$\pm$0.15 & 36.21$\pm$0.23 & 32.50$\pm$0.27 \\
            \cmidrule{2-5}
            & Rand & \textbf{65.42}$\pm$0.18 & \textbf{28.30}$\pm$7.31 & \textbf{24.42}$\pm$0.27 \\
            & DM & 63.76$\pm$0.38 & 12.61$\pm$0.46 & 8.50$\pm$0.89 \\
            \bottomrule
            \end{tabular}
        \end{sc}
    \end{center}
    \vskip -0.5cm
\end{table*}

\newpage
\subsection{Downstream performance of MCS w.r.t. $\ell_{\infty}$-norm} \label{app:linf-MCS}
\begin{figure}[ht]
    \vskip -0.2in
    \begin{center}
        \centerline{\includegraphics[width=\textwidth]{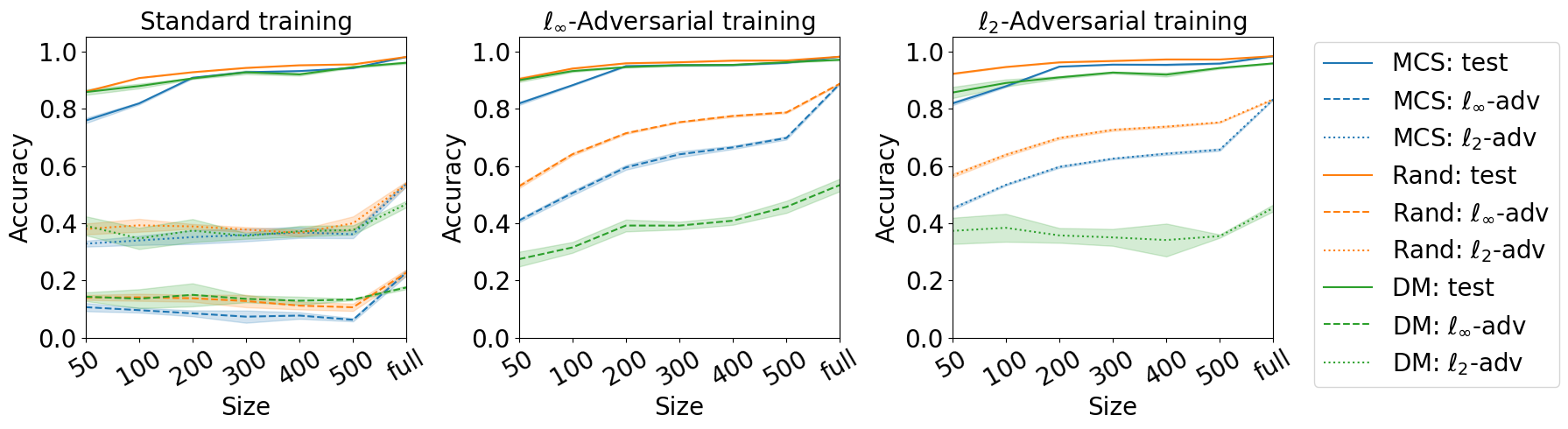}}
        \caption{Performance of standard and robust models trained over different compressed dataset from MNIST. The figures from left to right are standard training, $\ell_{\infty}$- and $\ell_2$- adversarial training. The blue, orange and green lines stand for \textbf{MCS}, \textbf{RAND}, and \textbf{DM} respectively. Here MCSs are obtained by solving \eqref{opt:eps-mfc} w.r.t. $\ell_{\infty}$-norm. The solid, dashed and dotted lines stand for test, $\ell_{\infty}$- and $\ell_2$- robust accuracy respectively. In the horizontal axis, ``full'' means the original dataset for \textbf{MCS} and \textbf{RAND}, and $k=5000$ for \textbf{DM}.} \label{fig:mnist-linf-results}
    \end{center}
    \vskip -0.5in
\end{figure}

\begin{figure}[ht]
    \vskip 0.2in
    \begin{center}
        \centerline{\includegraphics[width=\textwidth]{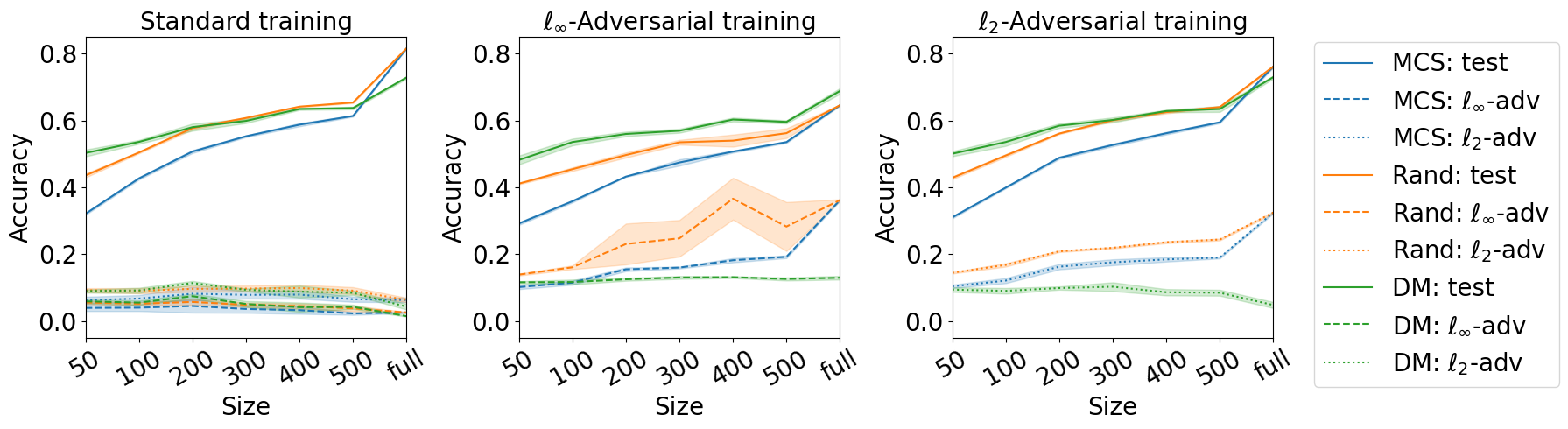}}
        \caption{Performance of standard and robust models trained over different compressed dataset from CIFAR10. The figures from left to right are standard training, $\ell_{\infty}$- and $\ell_2$- adversarial training. The blue, orange and green lines stand for \textbf{MCS}, \textbf{RAND}, and \textbf{DM} respectively. Here MCSs are obtained by solving \eqref{opt:eps-mfc} w.r.t. $\ell_{\infty}$-norm. The solid, dashed and dotted lines stand for test, $\ell_{\infty}$- and $\ell_2$- robust accuracy respectively. In the horizontal axis, ``full'' means the original dataset for \textbf{MCS} and \textbf{RAND}, and $k=4000$ for \textbf{DM}.} \label{fig:cifar10-linf-results}
    \end{center}
    \vskip -0.5in
\end{figure}

\newpage
\subsection{Comparison with classical and generalized adversarial training over MCS w.r.t. $\ell_{\infty}$-norm} \label{sec:compa-linf}
\begin{figure}[ht]
    \vskip -0.2in
    \begin{center}
        \centerline{\includegraphics[width=\textwidth]{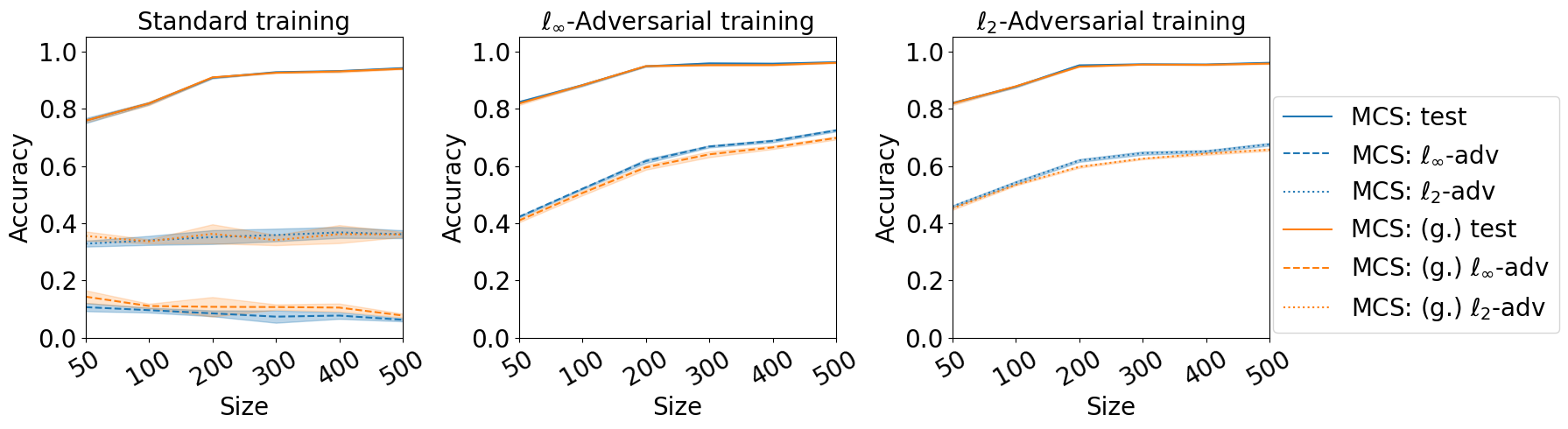}}
        \caption{Comparison of classical and generalized adversarial training over MCS of MNIST dataset. Here MCSs are obtained by solving \eqref{opt:eps-mfc} w.r.t. $\ell_{\infty}$-norm. The figures from left to right are standard training, $\ell_{\infty}$- and $\ell_2$- adversarial training. The blue and orange lines stand for classical and generalized training respectively. The solid, dashed and dotted lines stand for test, $\ell_{\infty}$- and $\ell_2$- robust accuracy respectively.} \label{fig:mnist-linf-comparison}
    \end{center}
    \vskip -0.5in
\end{figure}

\begin{figure}[ht]
    \vskip -0.2in
    \begin{center}
        \centerline{\includegraphics[width=\textwidth]{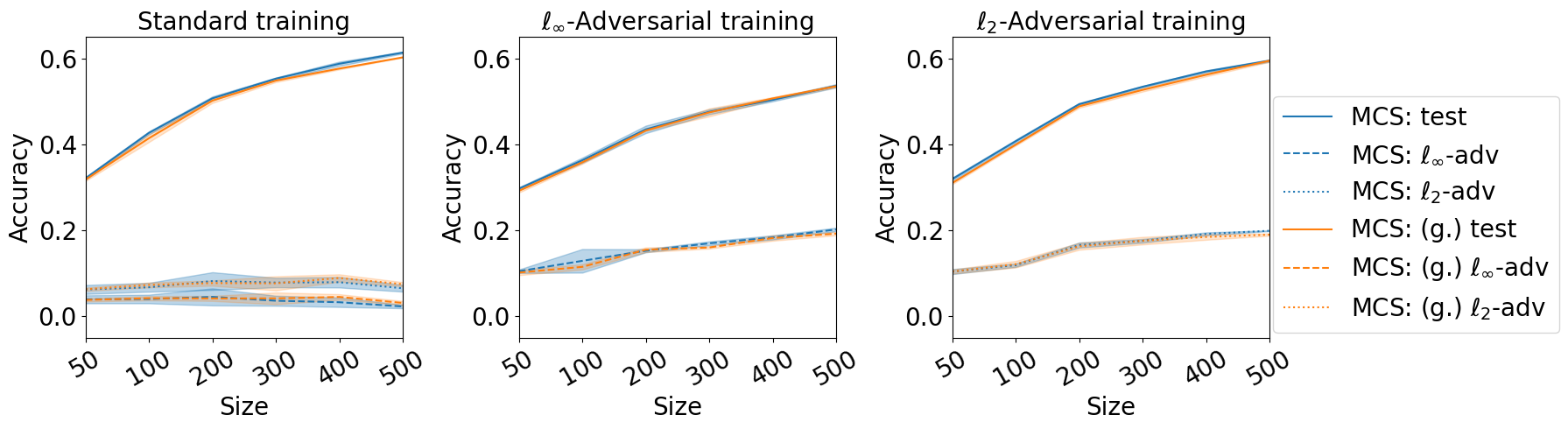}}
        \caption{Comparison of classical and generalized adversarial training over MCS of CIFAR10 dataset. Here MCSs are obtained by solving \eqref{opt:eps-mfc} w.r.t. $\ell_{\infty}$-norm. The figures from left to right are standard training, $\ell_{\infty}$- and $\ell_2$- adversarial training. The blue and orange lines stand for classical and generalized training respectively. The solid, dashed and dotted lines stand for test, $\ell_{\infty}$- and $\ell_2$- robust accuracy respectively.} \label{fig:cifar10-linf-comparison}
    \end{center}
    \vskip -0.5in
\end{figure}

\newpage
\subsection{Comparison with classical and generalized adversarial training over MCS w.r.t. $\ell_2$-norm} \label{app:compa-l2}
\begin{figure}[ht]
    \vskip -0.2in
    \begin{center}
        \centerline{\includegraphics[width=\textwidth]{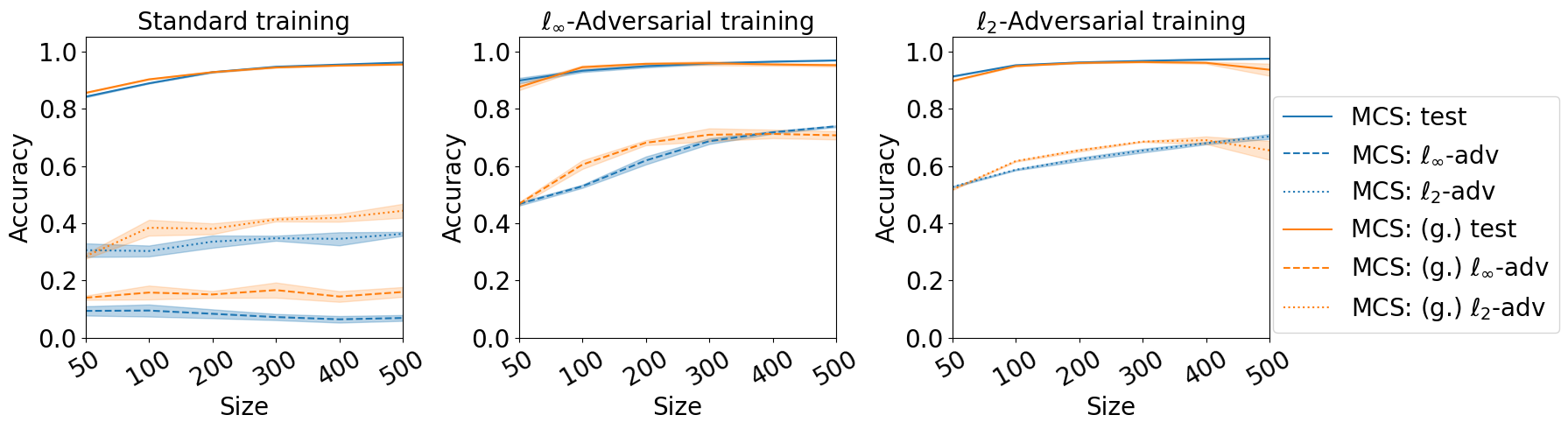}}
        \caption{Comparison of classical and generalized adversarial training over MCS of MNIST dataset. Here MCSs are obtained by solving \eqref{opt:eps-mfc} w.r.t. $\ell_{2}$-norm. The figures from left to right are standard training, $\ell_{\infty}$- and $\ell_2$- adversarial training. The blue and orange lines stand for classical and generalized training respectively. The solid, dashed and dotted lines stand for test, $\ell_{\infty}$- and $\ell_2$- robust accuracy respectively.} \label{fig:mnist-l2-comparison}
    \end{center}
    \vskip -0.5in
\end{figure}

\begin{figure}[ht]
    \vskip -0.2in
    \begin{center}
        \centerline{\includegraphics[width=\textwidth]{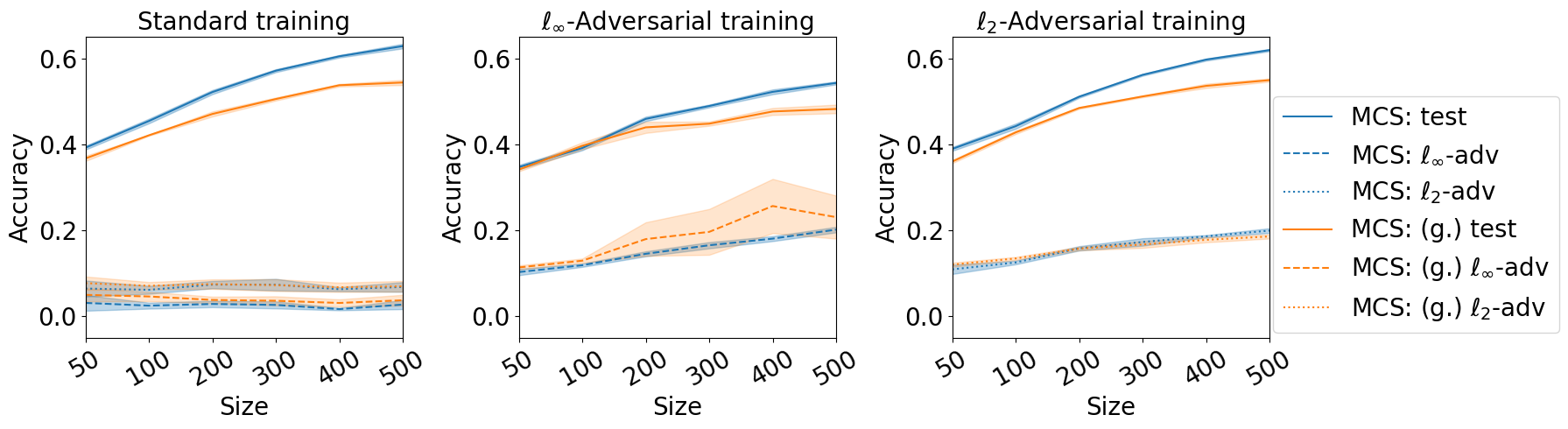}}
        \caption{Comparison of classical and generalized adversarial training over MCS of CIFAR10 dataset. Here MCSs are obtained by solving \eqref{opt:eps-mfc} w.r.t. $\ell_{2}$-norm. The figures from left to right are standard training, $\ell_{\infty}$- and $\ell_2$- adversarial training. The blue and orange lines stand for classical and generalized training respectively. The solid, dashed and dotted lines stand for test, $\ell_{\infty}$- and $\ell_2$- robust accuracy respectively.} \label{fig:cifar10-l2-comparison}
    \end{center}
    \vskip -0.5in
\end{figure}
\end{document}